# Comparing methods for Twitter Sentiment Analysis

Evangelos Psomakelis
Dept. of Informatics and Telematics
Harokopio University of Athens
9, Omirou Str.,
17778, Tavros, Greece
+302109549413
vpsomak@hua.gr

Konstantinos Tserpes
Dept of Informatics and Telematics
Harokopio University of Athens
9, Omirou Str.,
17778, Tavros, Greece
+302109549413
tserpes@hua.gr

Dimosthenis Anagnostopoulos
Dept of Informatics and Telematics
Harokopio University of Athens
9, Omirou Str.,
17778, Tavros, Greece
+302109549425
dimosthe@hua.gr

Theodora Varvarigou
Dept of Electrical and Computer Engineering
National Technical University of Athens,
9, Heroon Polytechniou Str, 15773, Zografos, Greece
+30 2107722484
dora@telecom.ntua.gr

## ABSTRACT
This work extends the set of works which deal with the popular problem of sentiment analysis in Twitter. It investigates the most popular document ("tweet") representation methods which feed sentiment evaluation mechanisms. In particular, we study the bag-of-words, n-grams and n-gram graphs approaches and for each of them we evaluate the performance of a lexicon-based and 7 learning-based classification algorithms (namely SVM, Naïve Bayesian Networks, Logistic Regression, Multilayer Perceptrons, Best-First Trees, Functional Trees and C4.5) as well as their combinations, using a set of 4451 manually annotated tweets. The results demonstrate the superiority of learning-based methods and in particular of n-gram graphs approaches for predicting the sentiment of tweets. They also show that the combinatory approach has impressive effects on n-grams, raising the confidence up to 83.15% on the 5-Grams, using majority vote and a balanced dataset (equal number of positive, negative and neutral tweets for training). In the n-gram graph cases the improvement was small to none, reaching 94.52% on the 4-gram graphs, using Orthodromic distance and a threshold of 0.001.

## Categories and Subject Descriptors
H.3.1 **Content Analysis and Indexing**

## General Terms
Algorithms

## 1. INTRODUCTION
The identification of sentiment in content from online social networks can be considered the "Holy Grail" for scientists in the Information Retrieval, Data Mining and Machine Learning fields. On one hand there is the challenge of the endeavour itself which often is a moving target: application requirements change (topics of interest, media, available resources, etc) and sentiment analysis solutions inherently cannot cope automatically with these changes. On the other hand there is the high demand, with a large number of organizations and individuals looking forward to lay their hands on mechanisms that will automatically harness the volume of data generated by users and assist them to evaluate public opinion regarding topics of interest (products, services, people, concepts, etc).

In this context, Twitter has comprised the most prominent playground for sentiment analysis solutions with businesses and scientists alike trying to tap into its users' enthusiasm for sharing opinions publicly online. It is not by chance that numerous works have suggested methods for implementing such mechanisms, e.g. [14], [8], [1], [20].

The most prominent of these methods rely on two approaches: lexicon-based and learning-based. In the first, the analysis of a document's expressed sentiment is achieved through its breakdown to words whose sentiment polarity is pre-defined in a lexicon. In learning-based, supervised classification algorithms are fed with pre-annotated (in terms of their sentiment polarity) documents, and are trained in order to autonomously classify future inputs.

In this work, we investigate some of the most well-used such methods based on both approaches. In the set of tests we include possible combinations of methods and report on their efficiency conducting experiments using a manually annotated Twitter dataset.

The major contributions of this work are: the extended comparison of sentiment polarity classification methods for Twitter text; the inclusion of combination of classifiers in the compared set, and; the aggregation and use of a number of manually annotated tweets for the evaluation of the methods. Especially regarding the latter, we consider it to be a main contribution in the sense that from past experience the automated annotation of tweets based on the detection of features like the emoticons ("☺", "☹", etc) has been problematic since it does not always reflect the case about the overall sentiment expressed by the author, especially when one considers the expression of no-sentiment ("neutral") through the text.

The rest of this report is structured as such: Section 2, defines the problem of sentiment analysis. Section 3 provides details about the representation models that are commonly met in the literature. Section 4, provides details about the experiments that were conducted and the results. Finally, Section 5, highlights the main conclusions from this work and reports on possible future directions for research and experimentation.

## 2. PROBLEM FORMULATION
work adopts and extends the definition of [16] for the sentiment polarity problem, according to which: "...*given an opinionated piece of text, wherein it is assumed that the overall opinion in it is about one single issue or item, classify the opinion as falling under one of two opposing sentiment polarities, or locate its position on the continuum between these two polarities.*". The latter allows room for defining three classes rather than the typical two (binary polarity problem). The third class refers to those text extracts that do not express either positive or negative sentiment, i.e. they are neutral.

In this context the problem of document-level sentiment analysis [4] is addressed. In this problem it is assumed that documents (in contrast to sentences or features) are opinionated regarding a particular topic. In the case of Twitter, the document is referred to as a "tweet" and it has a very specific form: a text message containing at most 140 characters.

The purpose is to create a program that will automatically identify whether the author of a tweet is expressing positive, negative or no sentiment about a topic.

The key challenge is to model the text in a way that the algorithm will use it as input and classify the text's sentiment polarity. Then, we need to identify or approximate the function that given the input modelled document, it will classify the document to the polarity class it belongs to. Formally and according to [2] it is: Given a collection of documents $D$ and the set of all classes $P_G = \{negative, neutral, positive\}$, the goal is to approximate the unknown target function $\Phi_G: D \to P_G$, which describes the polarization of documents according to a golden standard, by means of a function $\Phi'_G: T \to P_G$ that is called the general polarity classifier.

The identification of the function $\Phi'_G$ involves the identification of the golden standard, i.e. a reference model document. The distance of the tweets in question from the golden standard defines the class in which they belong. Hence, a third challenge is the definition of "distance".

An overview of the related work in representation models and classification algorithms that influenced this research is presented in the next Section.

## 3. RELATED WORK

One of the inherent difficulties when researching Sentiment Analysis problems is the translation of the textual data into a format that the computer can understand and process. For that exact function a number of Natural Language Processing (NLP) methods have been developed over the years. In this paper we will be using three of the most popular ones, the Bag of Words, the N-Grams and the N-Gram Graphs.

The Bag of Words [9] may be classified as the simplest method. According to this approach, the sentences of the document for which the machine needs to judge the sentiment it expresses, are split into a set of words using the space or the punctuation characters. These words form a virtual bag of words due to the fact that we don't keep any data indicating their ordering or their connection with their neighbors. Usually this method is assisted by a dictionary that correlates each word with a numerical value, showing its sentiment polarity [22]. The lack of contextual information though makes this correlation inaccurate in the case of "thwarted expectations" as explained by both [15] and [21]. This case is pretty common in reviews and can confuse every bag of words algorithm.

The N-Grams are pretty similar to the bag of words with one big difference; The text is split in pseudowords of equal length [16]. The length N is depended by the nature of the input documents and the problem at hand. Commonly, 2-Grams, 3-Grams and 4-Grams are used and other variations are largely rare [5, 6]. After the split these pseudowords form a bag similar to the one formed in the bag of words method. Using a dictionary in this case is not useful, because only a small -and in many cases random- set of pseudowords correspond to real words [2]. A more customized way of assigning sentiment values to each N-Gram is required. More details are provided in Section 4.

The popularity of social media stretched the two abovementioned methods to their limits. Documents to be analyzed became short, containing many abbreviations and neologisms, as well as many syntax and grammar errors. N-Gram graphs were suggested as an alternative. The formation of N-Grams remains at the core of their concept however, each one of them is depicted as a node in a graph. The graph denotes the position of each N-Gram in the sentence and its relation with its neighbors [7]. As such, each document is represented by a graph which in turn contains a number of nodes, each of which represents an N-Gram, a set of edges representing the neighboring of two N-Grams and a weight on each edge, representing the frequency with which this edge is encountered in the analyzed text. This frequency shows us how dominant this relation is in the analyzed text.

A graph can be assigned to a sentiment polarity class based on its comparison with the golden standard. In this case, this standard is a merged graph for each one of the three categories. The merged graph contains all the N-Grams of the individual training graphs, all their edges and an average weight on each edge [2]. This way each graph can be converted to a numerical vector, showing the distance from the three merged graphs. The shortest distance dictates the class to which the graph belongs.

The next step is to analyze the preprocessed dataset using various machine learning algorithms (classifiers) and categorization tools. In this paper Support Vector Machines (SVMs), Naïve Bayesian Networks, C4.5, Functional Trees, Best-First Trees, Multilayer Perceptrons and Logistic Regression algorithms are examined; in isolation but also in combinational experiments.

In general, Naïve Bayesian Networks, Functional Trees, Best-First Trees and the C4.5 algorithms try to create a categorization tree by dividing the possible values of each tested variable. Their difference is in the way they are creating the trees. For example C4.5, which is based on the ID3 algorithm, is using the Information Gain theory, which in turn is based on the Entropy theory, in order to perform the less possible splits, creating the smallest possible tree [17]. On the other hand, Naïve Bayesian Networks use a probabilistic model in order to maximize their accuracy with no consideration to the size of the produced tree [12]. Best-First Trees decides the "best" point to split the tree by a more arbitrary function, specified by the user in order to minimize the impurity between each new split [19].

Furthermore, Logistic Regression and Support Vector Machines try to find a mathematical function that can predict the correlation between the variables and the class. The Logistic Regression is using various logistic functions in order to calculate a function with a graphical representation that has the minimum distance from each of the training data points in the multidimensional space [18]. The Support Vector Machines use other mathematical functions, such as the minimum square function, in order to create another function with a similar graphical representation [13].

Finally, the Multilayer Perceptrons are an implementation of artificial neural networks. They consist of a number of nodes, grouped in different fully interconnected layers [11]. Each node is mapping its input values into a set of output values, using an internal function. This way each variable will be processed by one or more of node trees. The outputs of the first layer will become the inputs of the second and so on until at least one node of each layer has been activated. That way each one of the variables can contribute into the final classification decision with an intelligently calculated weight.

From a high level perspective, a similar work has been conducted and reported in [10]. The authors tested and compared a number of complete sentiment analysis methods, i.e. specific instances of NLP methods and algorithms suggested in the literature. In the present research the comparison emphasis was given to the combination of different NLP methods and machine learning algorithms detaching from one another. Some of these combinations were consistent with standardized sentiment analysis techniques but most of them were created just for testing purposes. Similar to the present research, Gonçalves et al. also tried to combine various methods but instead of creating a method by combining each algorithm's decisions in an environment with no a priori knowledge, they applied weights on the decision of each technique, based on a number of statistical values representing each techniques credibility.

In what follows we provide an analysis of the various experiments that we conducted in order to identify the best combination of representation model, classification algorithm and distance metric.

## 4. EXPERIMENTS

For the experiments we ran, we used a dataset consisting of 4451 tweets, assembled by various datasets that exist on the web. These tweets were rated manually by a number of researchers, according their sentiment polarity towards their subject. That way each tweet was assigned to a positive, neutral or negative category, helping us train the machine learning algorithms we used. After this categorization there were 1203 positive tweets, 1313 neutral tweets and 1935 negative tweets. This slight tendency towards negativity affected some of the algorithms, either causing increased or decreased accuracy.

In order to increase the effectiveness of the categorization on a later phase of the experimentation, the tweets had to be preprocessed. On this phase the preprocessing consisted of removing special characters that added no value to the sentiment polarity, such as the '#' character. Then the whole text of every tweet was converted to lower case characters and every web address in it was replaced by the keyword URL, since the actual link was of no importance, the important fact was that there was a link. As a last step the references to other users, using the '@' character, were replaced by the REF keyword since the username of the referred user had no impact on the sentiment polarity of the tweet. For example the tweet: "@Elli Expert settles for biofuel *Says it is efficient, ecofriendly -... http://t.co/aW14eUJJFH" was converted to "REF expert settles for biofuel says it is efficient, ecofriendly -... URL".

To compare the effectiveness of each experiment we compared just the confidence ratio of the categorization. We also filtered out the experiments that required unrealistic execution times or computational resources. Each experiment consisted of a NLP method that translated the textual data in a format more easily processable by the machine learning algorithms and one or more machine learning algorithms that were used to categorize the textual data. The categories were three: positive, neutral and negative, depending on the opinion that the author of the text was expressing about his or her subject.

Depending on the NLP method used, the experiments were split into three groups: the Bag of Words, N-Grams and N-Gram Graphs experiments. Then, 7 categorization algorithms were applied on the dataset formed by each NLP method, namely Support Vector Machine, Naïve Bayesian Networks, Logistic Regression, Multilayer Perceptrons, Best-First Trees, Functional Trees and C4.5. In a fourth experiment family those 7 algorithms were combined, forming a cooperative decision scheme with various improvements on their confidence rates, as it is presented later on.

Starting with the Bag of Words experiments, each tweet's text is split into a collection of words. This split did not take into consideration the relative position of each word so only the presence or absence of a word is examined; not its role in the sentence or its connection with neighboring words. To calculate each word's sentiment polarity the SentiWordNet 3.0 [3] was employed. SentiWordNet is a dictionary in which each word is associated to three numerical scores. The values of the scores indicate correspondingly how positive, negative, and "objective" (i.e., neutral) the word is.

Therefore, in this case, each tweet is modeled by a triplet of values in [0.0, 1.0]. Each value is the aggregate of the corresponding positive, negative and neutral scores of the words contained in the tweet. To classify the tweet in a single sentiment polarity the average of the three values is considered. By setting a positive and a negative threshold (gold standard) we were able to categorize a tweet as positive, neutral or negative. If for example the positive threshold was 0.5 and the negative was -0.5 then a tweet with average polarity -0.3 would be considered neutral, but a tweet with average polarity 0.6 would be considered positive. The thresholds in this case were set by trial and error.

Due to the particularities of tweets, many users are forced to be creative when it comes to bringing across their message without using many words. It is often the case that new words are created which are shorter than the words with a similar meaning. Thus the dictionary is proved to be an inaccurate solution in distinguishing the polarity of each tweet. This is made obvious when examining the confidence ratios that resulted by these experiments. In most cases the results were slightly above the random guess threshold of 33%. In some cases we noticed a slight increase up to 35% but it can be considered a statistical anomaly.

To improve this ratio a number of machine learning algorithms were introduced in the method. Those algorithms were used to automatically generate the positive and negative thresholds, reducing the number of false categorizations. A 10-fold cross validation procedure was used in order to train and test each algorithm in the same dataset. The resulting ratios were apparently higher but they still didn't manage to pass the 50% mark. The best result that was gathered was at 45.07% successful categorization using Naïve Bayesian Networks.

The next step was to change the NLP model. In this case N-Grams were used. The experiments were conducted for 3-Grams, 4-Grams and 5-Grams. The N-Grams method splits the text in pseudowords consisting of N characters each. The higher the N the more resources required to process the resulting dataset, up to a limit set by the length of the original text. After the split, the resulting pseudowords form a bag of pseudowords similar to the bag of words that was presented previously. In this case each word cannot be replaced by a sentiment value because each tweet is producing its own, unique N-Grams, and their unique nature prevents any dictionary from being effective. Thus an extra level of processing is needed in order to assign a numerical value to each N-Gram, showing its sentiment polarity.

This process for achieving so includes the counting of the number of times that each N-Gram appears in the positive, neutral or negative tweets and uses that frequency as its value. That way we have three values for each N-Gram, a positive, negative and a neutral frequency. We choose to consider only the greatest of the three frequencies during the categorization, ignoring the other two, in order to remain as close as we can to the bag of words model, which relates each word with only one numerical value. After that process we can replace each N-Gram with a numerical value, depending on its dominant category. If it is a positive N-Gram then it is replaced by its frequency, if it is a neutral N-Gram it is replaced by 0 and if it is a negative N-Gram then it is replaced by its frequency multiplied by -1. After that the procedure is exactly the same as the bag of words, i.e. the average of the numerical values is estimated and thresholds are set by the classifiers.

Due to the extra processing layer and the independence from dictionaries and existing words, one may notice a significant increase in the confidence rates. An important factor here is the length of the N-Grams, noticing there was an improvement in the confidence ratios as the window N length was raised from 3 to 4 and then 5 characters. The confidence rates in the case of 5-Grams are almost the double of a random prediction. The most successful results during these experiments were:

- 52.19% on 3-Grams, using logistic regression and a balanced dataset, limiting the data used to 3609 tweets, evenly distributed among the three categories.
- 65.21% on 4-Grams, using again logistic regression and the balanced dataset.
- 75.88% on 5-Grams, using logistic regression and the balanced dataset.

The third NLP method that was tested was the N-Gram Graphs. These graphs use the same N-Grams that were used in the previous experiment family but now they take into consideration the interactions that each N-Gram has with its neighbors. This places the individual N-Grams in a context, making their polarity value more foolproof than the rest. The tradeoff for this extra information is the increased computational cost.

In order to categorize each tweet, we create three merged graphs, one positive, one neutral and one negative (golden standards). These merged graphs contain all the graphs produced by the training data. Each one contains all the nodes and edges of the individual ones and averaged weights for each edge. Each new graph that is tested is compared with each one of these three graphs, using three different similarity functions [2]. That way a vector of nine numerical values, three for each merged graph, was produced for each tested tweet. These vectors were processed by the classifiers in order to set the appropriate threshold for categorization.

The resulting merged graphs though were unrealistically big, making it almost impossible to process them. To reduce their volume, the graphs were pruned, by setting an arbitrary pruning threshold on the edge weights. Experiments were conducted with two different thresholds in order to distinguish the effect that the increase or decrease of the threshold has on the resulting confidence ratio. According to the results that will be presented shortly, the pruning threshold greatly affects the resulting confidence ratio. Our experiments with a threshold equal to 0.01 reached a limit of 67.12% using logistic regression. When the threshold was reduced to 0.001 we got a maximum ratio of 94.53% using multilayer perceptrons.

Following these results a fourth family of experiments was implemented, the Combinational experiments. During these tests all seven algorithms were employed in a cooperative decision model. Several ways of making a categorization decision were studied, taking into consideration the opinion of every algorithm. Those methods consisted of a simple majority vote, an average opinion and several distance functions examining how far the collective opinion was from the centroid of a positive, neutral and negative opinion cluster. These distance functions were the Euclidian distance, the Manhattan distance, the cosine dissimilarity, the Orthodromic distance and the Chebychev distance (Hertz, 2006). The results were encouraging in the N-Gram experiments, raising the confidence up to 83.15% on the 5-Grams, using majority vote and the balanced dataset. In the N-Gram Graph cases the improvement was small to none, reaching 94.52% on the 4-Gram Graphs, using Orthodromic distance and a threshold of 0.001.

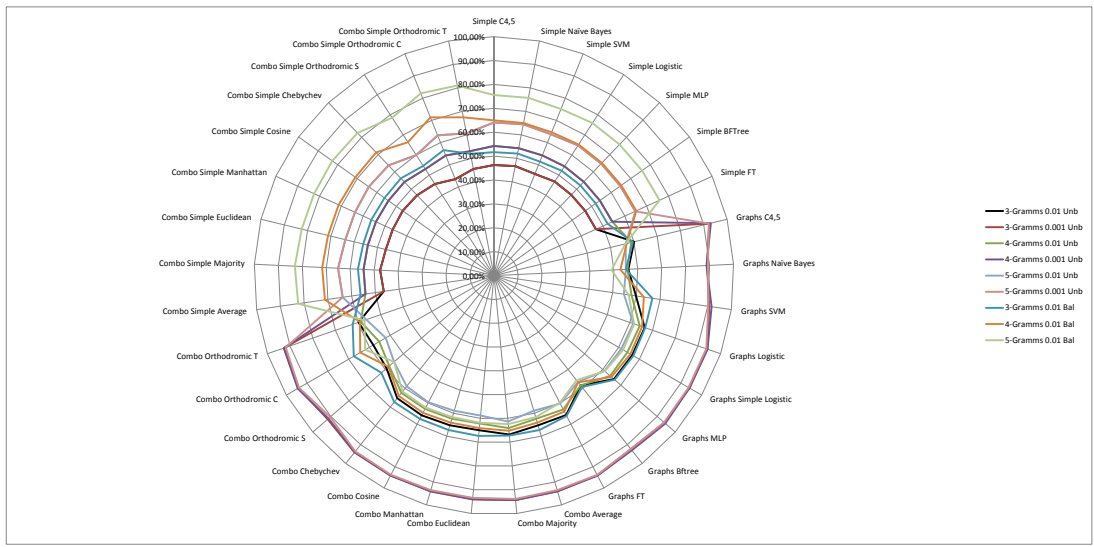

**Figure 1: Summary presentation of the performance of the various methods**

Figure 1 depicts the collection of resulting confidence rates. The center of the circle represents 0% success and the outer circle 100% success. To compose this graph the results from more than 250 experiments were used. The bottom hemisphere shows the results of the N-Gram Graph experiments, both the simple and the combinational ones. According to these results, the 0.001 threshold experiments are surpassing very clearly all other experiments. Instead the 0.01 threshold ones are about on the same level as the N-Gram experiments of the upper hemisphere.

## 5. CONCLUSIONS

In this work we presented an analysis and overview of the most prominent methods for sentiment analysis in Twitter. The emphasis was put on the various NLP models and the combinations of various classifiers. Lexicon-based methods were also used.

The results demonstrated the superiority of n-gram graphs in capturing the expressed sentiment in a document and specifically in tweets. They also demonstrated the improvements that various combinations of NLP methods and machine learning algorithms can induce in the confidence rates of some sentiment analysis techniques.

The innovation of this work is concentrated in the meticulous evaluation of the efficiency of various sentiment analysis mechanisms using manually annotated datasets, as well as in the demonstration of the possibility to combine methods, creating new techniques for enhancing the quality of the outcome.

## 6. ACKNOWLEDGMENTS

This work has been supported by the Consensus project (http://www.consensus-project.eu) and has been partly funded by the EU Seventh Framework Programme, theme ICT-2013.5.4: ICT for Governance and Policy Modelling under Contract No. 611688.